%% file: main.tex
\documentclass[11pt,a4paper]{article}
\usepackage[hyperref]{naaclhlt2018}
\usepackage{times}
\usepackage{latexsym}
\usepackage{algorithm}
\usepackage{algorithmic}
\usepackage{amsmath}
\usepackage{amsfonts}
\usepackage{mathtools}
\usepackage{dsfont}
\usepackage{flushend}
\usepackage{url}
\usepackage{multirow}

\usepackage{times}
\usepackage{graphicx} 
\usepackage{subfigure}

\usepackage{natbib}

\usepackage{hyperref}

\aclfinalcopy 

\newcommand{\task}{\mathcal{T}}
\newcommand{\loss}{\mathcal{L}}
\newcommand{\inp}{\mathbf{x}}
\newcommand{\learner}{f}
\newcommand{\lossi}{\loss_{\task_i}}
\newcommand{\target}{\mathbf{y}}

\include{defs}

\title{Natural Language to Structured Query Generation via Meta-Learning}

\author{Po-Sen Huang$^\star$, Chenglong Wang$^{\dagger}$, Rishabh Singh$^{\star}$, Wen-tau Yih$^{\ddagger}$, Xiaodong He\thanks{\hspace{.06in}Work performed while XH was at Microsoft Research.}\hspace{0.015in}$^{~\diamond}$ \\
	$^\star$Microsoft Research, $^\dagger$University of Washington,\\
	$^\ddagger$Allen Institute for Artificial Intelligence, $^\diamond$JD AI Research\\
	{\small\texttt{pshuang@microsoft.com,~clwang@cs.washington.edu,}}\\ {\small\texttt{rishabh.iit@gmail.com, scottyih@allenai.org, xiaodong.he@jd.com}}
	\\
} 
\date{}

\begin{document}
\maketitle
\begin{abstract}
In conventional supervised training, a model is trained to fit all the training examples.
However, having a monolithic model may not always be the best strategy, as examples could 
vary widely.
In this work, we explore a different learning protocol that treats each example as a 
unique \emph{pseudo-task}, by reducing the original learning problem to 
a few-shot meta-learning scenario with the help of a domain-dependent \emph{relevance function}.\footnote{The source code is available at \url{https://github.com/Microsoft/PointerSQL}.} 
When evaluated on the \mbox{WikiSQL} dataset, our
approach leads to faster convergence and achieves 1.1\%--5.4\% absolute accuracy gains over the non-meta-learning counterparts. 
\end{abstract}
 
%

\input{intro.tex}

\input{background-MAML.tex}
\input{approach.tex}
\input{exp.tex}


\section{Related Work}
{\bf Meta Learning}
One popular direction of meta-learning \citep{Thrun:1998, schmidhuber1987, naik1992meta} is to train a meta-learner that learns how to update the parameters of the learner’s model (a lower level model) \citep{bengio1992optimization,  schmidhuber1992learning}.
This direction has been applied to learning to optimize deep neural networks \citep{hochreiter2001learning, andrychowicz2016learning, li2017learning, ha2016hypernetworks}.
Few-shot learning methods have also applied meta-learning approaches to image recognition \citep{koch2015siamese, ravi2016optimization, vinyals2016matching} and reinforcement learning \citep{MAML}.
Given that the few-shot learning setup cannot directly work in standard supervised learning problems, we explore reducing a regular supervised learning problem to the few-shot meta-learning scenario by creating {pseudo-tasks} with a {relevance function}. 

{\bf Semantic Parsing}
Mapping natural language to logic forms has been actively studied in natural language processing research \citep{Zettlemoyer:2005, Giordani_Moschitti2010, Artzi:2011, berant2013semantic, vlachos2014new, yih2014semantic, yih2015semantic, wang2015building, GolubH16, iyer2017learning, krishnamurthy2017neural}.
However, unlike conventional approaches, which fit one model for all training examples, the proposed approach learns to adapt to new tasks.
By using the support set based on the relevance function, the proposed model can adapt to a unique model for each example.

{\bf Program Induction / Synthesis}
Program induction \citep{reed2015neural, neelakantan2015adding, graves2014neural, yin2015neural, npi} aims to infer latent programs given input/output examples, while program synthesis models \citep{wikisql, parisotto2016neuro} aim to generate explicit programs and then execute them to get output.
The learner model we used in this work follows the line of program synthesis models and trains on pairs of natural language (question) and program (SQL) directly. 


 \vspace{-1mm}
\section{Conclusion}
 \vspace{-1mm}
In this paper, we propose a new learning protocol that reduces a regular supervised learning problem to the few-shot meta-learning scenario.
This is done by effectively creating \emph{pseudo-tasks} with the help of a \emph{relevance function}. 
When evaluated on the newly released, large semantic parsing dataset, \mbox{WikiSQL}, our approach leads to faster convergence and enjoys 1.1\%--5.4\% absolute accuracy gains over the non-meta-learning counterparts, achieving a new state-of-the-art result.


While the initial finding is encouraging, we believe the potential of this meta-learning framework has not yet
been fully realized. In the future, we plan to explore more variations of the meta-learning setup, such
as using different relevance functions, including the ones that are jointly learned.  We also would like to
understand this approach better by testing it on more natural language processing tasks.

 \vspace{-1mm}
\section*{Acknowledgments}
 \vspace{-1mm}
We thank Chelsea Finn and Eugene Brevdo for helpful discussions in meta-learning, and Adith Swaminathan, Asli Celikyilmaz, and anonymous reviewers for their valuable feedback. 


\bibliography{naaclhlt2018}
\bibliographystyle{acl_natbib}

\input{appendix}

\end{document}

%% file: defs.tex
\newcommand{\ignore}[1]{}
\newcommand{\eat}[1]{}

\long\def\ignorethis#1{}



%% file: intro.tex
\section{Introduction}
\label{sec:intro}

Conventional supervised training is a pervasive paradigm for NLP problems. In this setting, 
a model is trained to fit all the training examples and their corresponding targets.
However, while sharing the same surface form of the prediction task, examples of the same problem may vary widely.
For instance, \emph{recognizing textual entailment} is a binary classification problem on whether the hypothesis follows
a given textual statement, but the challenge datasets consist of a huge variety of inference categories and genres~\cite{Dagan13,Williams2017ABC}. 
Similarly, for a semantic parsing problem that maps natural language questions to
SQL statements, the number of conditions in a SQL query or the length of a question can vary substantially~\cite{wikisql}.

The inherently high variety of the examples suggests an alternative training protocol: instead of learning a monolithic, one-size-fits-all model, it could be more effective to learn multiple models, where each one is designed for a specific ``task" that covers a group of \emph{similar} examples.  However, this strategy is faced with at least two difficulties. As the number of tasks increases, each task will have much fewer training examples for learning a robust model. In addition, the notion of ``task", namely the group of examples, is typically not available in the dataset.  

\begin{figure}[t]
	\centering
	\includegraphics[width=0.9\columnwidth]{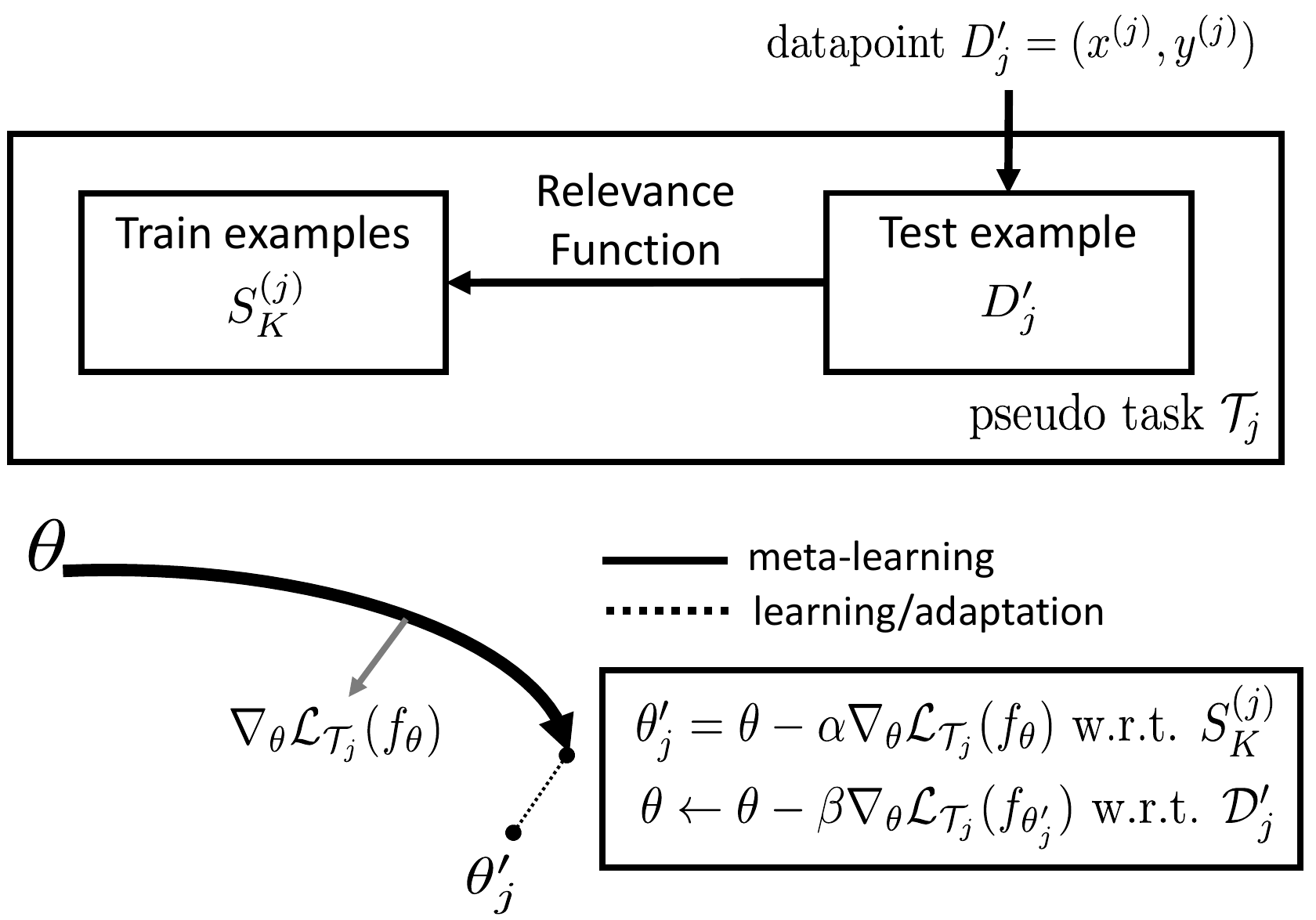}
	\vspace{-5mm}
	\caption{{\small Diagram of the proposed framework. (Upper) we propose using a relevant function to find a support set $S_K^{(j)}$ from all training datapoints given a datapoint $D_j'$ for constructing a pseudo-task $\mathcal{T}_j$ as in the few-shot meta-learning setup.  
	(Bottom) We optimize the model parameters $\theta$ such that the model can learn to adapt a new task with parameters $\theta_j'$ via a few gradient steps on the training examples of the new task. The model is updated by considering the test error on the test example of the new task. See Section~\ref{sec:background} for detail.}}
	\label{fig:overview_diagram}
\end{figure}

\vspace{-0.5mm}
In this work, we explore this alternative learning setting and address the two difficulties by adapting the \emph{meta-learning} framework.  Motivated by the few-shot learning scenario~\cite{andrychowicz2016learning, ravi2016optimization, vinyals2016matching}, meta-learning aims to learn a general model that 
can quickly adapt to a new task given very few examples without retraining the model from scratch~\cite{MAML}.
%
We extend this framework by effectively creating \emph{pseudo-tasks} with the help of a \emph{relevance function}. 
During training, each example is viewed as the test example of an individual ``task'', 
where its top-$K$ relevant instances are used as training examples for this specific task. 
A general model is trained for all tasks in aggregation. 
Similarly during testing, instead of applying the general model directly, the top-$K$ relevant instances (in the
training set) to the given test example are first selected to update the general model, which then makes
the final prediction.
The overview of the proposed framework is shown in Figure \ref{fig:overview_diagram}.

\vspace{-0.5mm}
When empirically evaluated on a recently proposed, large semantic parsing dataset, \mbox{WikiSQL}~\cite{wikisql}, our
approach leads to faster convergence and achieves 1.1\%--5.4\% absolute accuracy gain over the non-meta-learning counterparts,
establishing a new state-of-the-art result.  More importantly, we demonstrate how to design a relevance function to
successfully reduce a regular supervised learning problem to a meta-learning problem.  To the best of our knowledge,
this is the first successful attempt in adapting meta-learning to a semantic task.

%% file: background-MAML.tex
\section{Background: Meta-Learning}
\label{sec:background}




Our work is built on the recently proposed Model-Agnostic Meta-Learning (MAML) framework~\citep{MAML}, which we describe briefly here.
MAML aims to learn the {\it learners} (for the tasks) and the {\it meta-learner} in the few-shot meta-learning setup~\citep{vinyals2016matching, andrychowicz2016learning, ravi2016optimization}.
Formally, it considers a model that is represented by a function $\learner_\theta$ with parameters $\theta$. 
When the model adapts to a new task $\task_i$, the model changes parameters $\theta$ to $\theta_i'$, 
where a task contains $K$ training examples and one or more test examples ($K$-shot learning).
MAML updates the parameters $\theta_i'$ by one or a few rounds of gradient descent based on the training examples of task $\task_i$.
For example, with one gradient update,
$$
\theta_i'=\theta-\alpha \nabla_\theta  \lossi(  \learner_\theta ),
$$
where the step size $\alpha$ is a hyper-parameter; 
$\lossi(  \learner_\theta )$ is a loss function that evaluates the error between the prediction $\learner_\theta(\inp^{(j)})$ and target $\target^{(j)}$, 
where $\inp^{(j)}, \target^{(j)}$ are an input/output pair sampled from the training examples of task $\task_i$. 
Model parameters $\theta$ are trained to optimize the performance of $\learner_{\theta_i'}$ on the unseen test examples from $\task_i$ across tasks.
The meta-objective is:
\begin{align*}
\min_\theta \sum_{\task_i \sim p(\task)}  \lossi ( \learner_{\theta_i'})
= \sum_{\task_i \sim p(\task)}  \lossi ( \learner_{\theta - \alpha \nabla_\theta \lossi(f_\theta)})
\end{align*}
The goal of MAML is to optimize the model parameters $\theta$ such that the model can learn to adapt new tasks with parameters $\theta_i'$ 
via a few gradient steps on the training examples of new tasks.
The model is improved by considering how the test error on unseen test data from $\task_i$ changes with respect to
the parameters. 

The meta-objective across tasks is optimized using stochastic gradient descent (SGD). The model parameters $\theta$ are updated as follows:
\[
\label{eq:metaupdate}
\theta \leftarrow \theta - \beta \nabla_\theta \sum_{\task_i \sim p(\task)}  \lossi ( \learner_{\theta_i'}), 
\]
where $\beta$ is the meta step size.

%% file: approach.tex
\section{Approach}
\label{sec:model}

\label{sec:problem_formulation}

As discussed in Section~\ref{sec:intro}, to reduce traditional supervised learning to a few-shot meta-learning problem, 
we introduce a \emph{relevance function}, which effectively helps group examples to form \emph{pseudo-tasks}.
Because the relevance function is problem-dependent, we first describe the semantic parsing problem below, followed by the design of our
relevance function and the complete algorithm.

\subsection{The Semantic Parsing Task}
\label{sec:wikisql_task}
The specific semantic parsing problem we study in this work 
is to map a natural language question
to a SQL query, which can be executed against a given table to find the answer to the original question.
In particular, we use the currently largest natural language questions to SQL dataset, \mbox{WikiSQL}~\citep{wikisql},
to develop our model and to conduct the experiments.

\ignore{
, which is the largest natural language to SQL dataset.
In particular, the input contains two parts: a natural language question stating the query for a table, and the SQL table itself.
The output is a SQL query that formalizes the intent of the natural language query.
}

\subsection{Relevance Function}
\label{sec:rel_function}


The intuition behind the design of a relevance function is that examples of the same type should have higher scores.
For the questions to SQL problem, we design a simple relevance function that depends on (1) the predicted type of
the corresponding SQL query and (2) the question length.

There are five SQL types in the \mbox{WikiSQL} dataset: \{\texttt{Count}, \texttt{Min}, \texttt{Max}, \texttt{Sum}, \texttt{Avg}, \texttt{Select}\}.
We train a SQL type classifier $f_{sql}$ using SVMs with bag-of-words features of the input question, which
achieves 93.5\% training accuracy and 88\% test accuracy in SQL type prediction.
Another soft indication on whether two questions can be viewed as belonging to the same ``task" is their lengths, as they correlate to
the lengths of the mapped SQL queries.  The length of a question is the number of tokens in it after normalizing entity mentions to single tokens.\footnote{Phrases in questions that can match some table cells are treated as entities.} 
Our relevance function only considers examples of the same predicted SQL types. If examples $\inp^{(i)}$ and $\inp^{(j)}$ have the same
SQL type, then their relevance score is $1-|{q_{len}}(\inp^{(i)})-q_{len}(\inp^{(j)})|$, where $q_{len}$ calculates the question length. 
Notice that the relevance function does not need to be highly accurate as there is no formal definition on which examples should be
grouped in the same pseudo-task.  A heuristic-based function that encodes some domain knowledge typically works well based on our preliminary
study.  In principle, the relevance function can also be jointly learned with the meta-learning model, which we leave for future work.

\subsection{Algorithm}

Given a relevance function, the adaptation of the meta-learning using the MAML framework can be summarized in Algorithm~\ref{alg:maml}, called Pseudo-Task MAML (PT-MAML).
For each training example $\inp^{(j)}$, we create a pseudo-task $\task_j$ using the top-$K$ relevant examples as the support set $\mathcal{S}^{(j)}_K$~(Step 1).
The remaining steps of the algorithm mimics the original MAML design, update task-level models~(Step 8) and the meta-level, general model~(Step 10) using gradient
descent.

\begin{algorithm}[t]
	\caption{Pseudo-Task MAML (PT-MAML)}
	\label{alg:maml}
	\begin{algorithmic}[1]
		\REQUIRE Training Datapoints  $\mathcal{D}=\{\inp^{(j)}, \target^{(j)}\}$
		\REQUIRE $\alpha$, $\beta$: step size hyperparameters
		\REQUIRE $K$: support set size hyperparameter
		\STATE Construct a task $\task_j$ with training examples using a support set $\mathcal{S}^{(j)}_K$ and a test example $\mathcal{D}_j'=(\inp^{(j)}, \target^{(j)})$.
		\STATE Denote $ p(\task)$ as distribution over tasks
		\STATE Randomly initialize $\theta$
		\WHILE{not done}
		\STATE Sample batch of tasks $\task_i \sim p(\task)$
		\FORALL{$\task_i$}
		\STATE Evaluate $\nabla_\theta \lossi(\learner_\theta)$ using $\mathcal{S}^{(j)}_K$
		\STATE Compute adapted parameters with gradient descent: $\theta_i'=\theta-\alpha \nabla_\theta  \lossi(  \learner_\theta )$
		\ENDFOR
		 \STATE Update $\theta \leftarrow \theta - \beta \nabla_\theta \sum_{\task_i \sim p(\task)}  \lossi ( \learner_{\theta_i'})$ using each $\mathcal{D}_i'$ from $\task_i$ and $\lossi$ for the meta-update
		\ENDWHILE
	\end{algorithmic}
\end{algorithm}

%% file: exp.tex
\section{Experiments}
\vspace{-1mm}
\label{sec:experiments}

In this section, we introduce the WikiSQL dataset and preprocessing steps, the learner model in our meta-learning setup, and the experimental results. 

\ignore{
We apply our approach to the WikiSQL dataset.  Details of the dataset and preprocessing steps can be found in Appendix A. 
Here we focus our discussion on the basic learner model and the results.
}
\ignore{
In this paper, we evaluate our approach on a recent proposed WikiSQL dataset \citep{wikisql}. The details of the dataset and preprocessing steps are introduced in Appendix~\ref{sec:dataset}. 
}

\vspace{-0.5mm}
\subsection{Dataset}
\label{sec:dataset}
\vspace{-0.5mm}
We evaluate our model on the WikiSQL dataset \citep{wikisql}.
We follow the data preprocessing in~\cite{chenglong}. 
Specifically, we first preprocess the dataset by running both tables and question-query pairs through Stanford Stanza \citep{manning2014stanford} using the script included with the WikiSQL dataset, which normalizes punctuations and cases of the dataset.
We further normalize each question based on its corresponding table: for table entries and columns occurring in questions or queries, we normalize their format to be consistent with the table.
After preprocessing, we filter the training set by removing pairs whose ground truth solution contains constants not mentioned in the question, as our model requires the constants to be copied from the question.
We train and tune our model only on the filtered training and filtered development set, but we report our evaluation on the full development and test sets. We obtain 59,845 (originally 61,297) training pairs, 8,928 (originally 9,145) development pairs and 17,283 test pairs (the test set is not filtered).

\vspace{-0.5mm}
\subsection{Learner Model}
\vspace{-0.5mm}
We use the model of \citet{chenglong} as the {\it learner} in our meta-learning setup. The model is a grammar-aware Seq2Seq encoder-decoder model with attention \citep{cho2014learning, bahdanau2014neural}.
The encoder is a bidirectional LSTM, which takes the concatenation of the table header (column names) of the queried table and the question as input to learn a joint representation.
%
The decoder is another LSTM with attention mechanism. 
There are three output layers corresponding to three decoding types, which restricts the vocabulary it can sample from at each decoding step.
The three decoding types are defined as follows: 

\begin{itemize}\itemsep-2pt
\vspace{-2mm}
	\item $\tau_V$ (SQL operator): The output has to be a SQL operator, i.e., a terminal from $V$ $=$ \{$\texttt{Select}$, $\texttt{From}$, $\texttt{Where}$, $\texttt{Id}$, $\texttt{Max}$, $\texttt{Min}$,	$\texttt{Count}$, $\texttt{Sum}$, $\texttt{Avg}$, $\texttt{And}$, $=$, $>$, $\geq$, $<$, $\leq$, $\texttt{<END>}$, $\texttt{<GO>}$\}.
\vspace{-1mm}
	\item $\tau_C$ (column name): The output has to be a column name, which will be copied from either the table header or the query section of the input sequence. Note that the column required for the correct SQL output may or may not be mentioned explicitly in the question.
\vspace{-1mm}
	\item $\tau_Q$ (constant value): The output is a constant that would be copied from the question section of the input sequence.
	\vspace{-1mm}
\end{itemize}

The grammar of SQL expressions in the the WikiSQL dataset can be described in regular expression as  ``\texttt{Select} $f$ $c$ \texttt{From} $t$ \texttt{Where} ($c$ $op$ $v$)$^*$'' ($f$ refers to an aggregation function, $c$ refers to a column name, $t$ refers to the table name, $op$ refers an comparator and $v$ refers to a value). 
The form can be represented by a decoding-type sequence  $\tau_V \tau_V \tau_C \tau_V \tau_C \tau_V (\tau_C \tau_V \tau_Q)^*$, which will ensure only decoding-type corrected tokens can be sampled at each decoding step.

\newcite{chenglong} propose three cross-entropy based loss functions: 
``Pointer loss'', which is the cross-entropy between target index and the chosen index, ``Max loss'', 
which computes the probability of copying a token $v$ in the input as the
maximum probability of pointers that point to token $v$, and ``Sum loss'', which computes the probability of copying a token $v$ in the input as the sum of probabilities of pointers that point to token $v$. 
See \cite{chenglong} for more detail. 

\setlength{\tabcolsep}{5pt}

\begin{table}[t]
	{\small
		\centering
		\begin{tabular}{|l|c|c|c|c|}
			\hline
			\multirow{ 2}{*}{Model}      & \multicolumn{2}{c|}{Dev} & \multicolumn{2}{c|}{Test} \\ \cline{2-5} 
			& $Acc_{lf}$    & $Acc_{ex}$   &$Acc_{lf}$    & $Acc_{ex}$  \\ \hline
			PointerNet   \shortcite{wikisql} & 44.1\%  & 53.8\%   & 43.3\%  & 53.3\%     \\ \hline
			Seq2SQL   \shortcite{wikisql} & 49.5\%  & 60.8\% & 48.3\% & 59.4\%      \\ \hline
			Pointer loss  \shortcite{chenglong}     &  46.8\%  &  52.1\%      &   46.1\%     &     51.8\%       \\ \hline
			Meta + Pointer loss  &  {\bf 52.0\%}  &  {\bf 57.7\%}  & {\bf 51.4\%}  & {\bf 57.2\%}   \\ \hline
			Max loss  \shortcite{chenglong}     &  61.3\%  &  66.9\%      &   60.5\%     &     65.8\%       \\ \hline
			Meta + Max loss  &  {\bf 62.1\%}  &  {\bf 67.3\%}  & {\bf 61.6\%}  & {\bf 67.0\%}   \\ \hline
			Sum loss  \shortcite{chenglong}     &  62.0\%  &  67.1\%      &   61.5\%     &     66.8\%       \\ \hline
			Meta + Sum loss  &  {\bf 63.1\%}  &  {\bf 68.3\%}  & {\bf 62.8\%}  & {\bf 68.0\%}   \\ \hline
		\end{tabular}
	} \vspace{-4mm}
		\caption{{Experimental Results on the WikiSQL dataset, where $Acc_{lf}$ represents the logical form accuracy and $Acc_{ex}$ represents the SQL execution accuracy. ``Pointer loss'', ``Max loss'', and ``Sum loss'' are the non-meta-learning counterpart from \citet{chenglong}.} ``Meta + X'' denotes the meta-learning model with learner ``X''. }
		\label{tab:exp_results}
\end{table}

\setlength{\tabcolsep}{6pt}

 \vspace{-1mm}

\vspace{-.5mm}
\subsection{Model Hyperparameters}
\vspace{-.5mm}

We use the pre-trained n-gram embeddings by \citet{hashimoto2016joint} (100 dimension) and the GloVe word embedding (100 dimension) by \citet{pennington2014glove}; each token is embedded into a 200 dimensional vector.
The encoder is a 3-layer bidirectional LSTM with hidden states of size 100, and the decoder is a 3-layer unidirectional LSTM with hidden states of size 100.
The model is trained with question-query pairs with a batch size of 200 for 100 epochs.
During training, we clip gradients at 5 and add gradient noise with $\eta=0.3$, $\gamma=0.55$ to stabilize training \citep{neelakantan2015adding}.
We found the meta-learning model is trained stably without back-propagating to second order gradients.
We select the support set size $K$ to be 2 based on the development set.
Empirically, the performance does not improve when we use a larger $K$.
We set the learning rates $\alpha=0.001$ and $\beta=0.1$ based on the development set.
The model is implemented in Tensorflow and trained using Adagrad~\citep{duchi2011adaptive}.

 \vspace{-.5mm}
\subsection{Results}
 \vspace{-.5mm}
Table~\ref{tab:exp_results} shows the experimental results of our model on the WikiSQL dataset.
We select the model based on the best logical form accuracy on the development set, and 
compare our results to augmented pointer network and the Seq2SQL model (with RL) in~\cite{wikisql}.
Both logical form accuracy (denoted by {\it $Acc_{lf}$}) that compares the exact SQL syntax match, and the SQL execution results (denoted by {\it $Acc_{ex}$}) are reported.
We compare our approach with its non-meta-learning counterpart using ``Pointer loss'', ``Max loss'', and ``Sum loss'' losses from \cite{chenglong}.
Our model achieves 1.1\%--5.3\% and 1.2\%--5.4\% gains on the test set logical form and execution accuracy, respectively.

\begin{figure}[t]
	\centering
	\includegraphics[width=0.73\columnwidth]{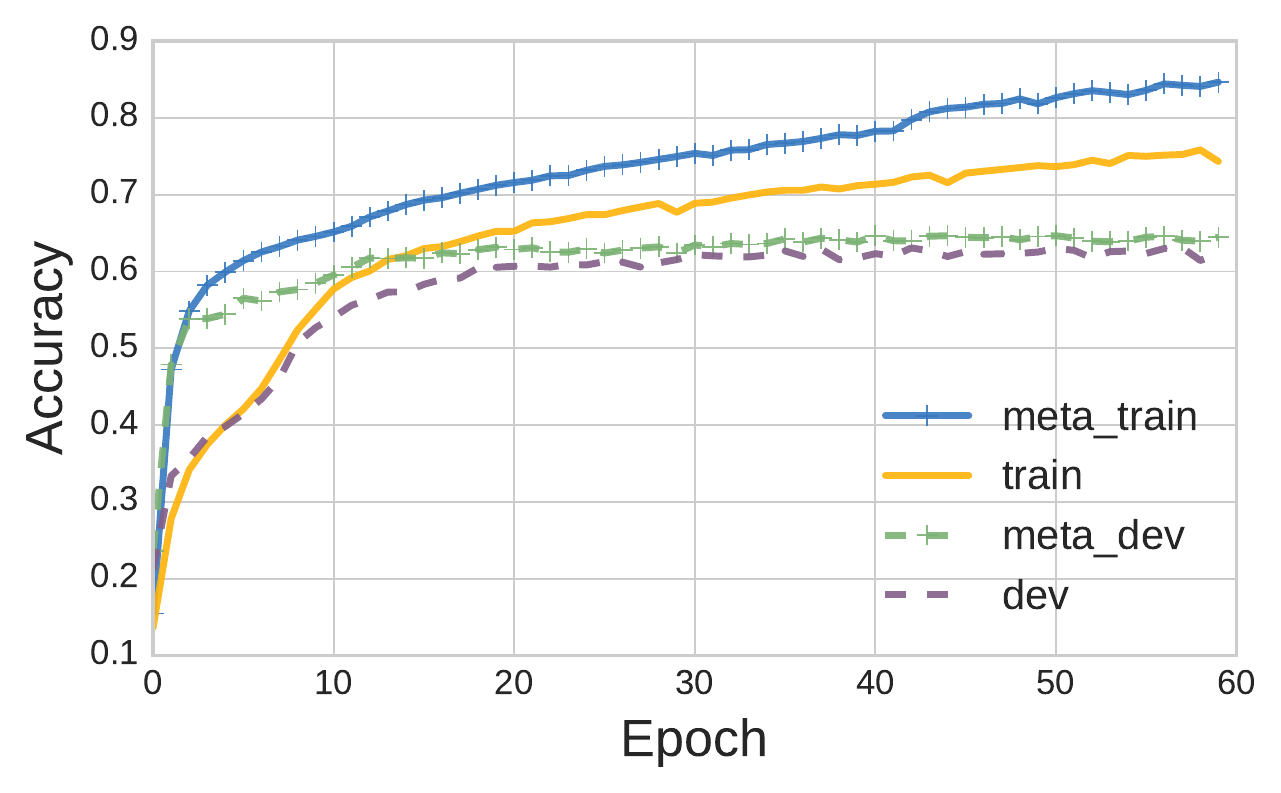}
	\vspace{-6mm}
	\caption{{Logical form accuracy comparison, where ``meta\_train'' and ``meta\_dev'' are the train and development set accuracy using the ``Meta + Sum loss'' model, ``train'' and ``dev'' are the train and development set accuracy using the ``Sum loss'' model \citep{chenglong}.}}
	 \vspace{-2mm}
		\label{fig:accuracy}
\end{figure}

We also investigate the training and development set logical form accuracy over different epochs by ``Meta + Sum loss'' and ``Sum loss'' models.
The results are shown in Figure~\ref{fig:accuracy}.
One interesting observation is that the ``Meta + Sum loss'' model converges much faster than the ``Sum loss'' model especially in the first 10 epochs.
We attribute this improvement to the ability to adapt to new tasks even with a small number of training examples.

\begin{figure}[t]
	\centering
	\includegraphics[width=0.9\columnwidth]{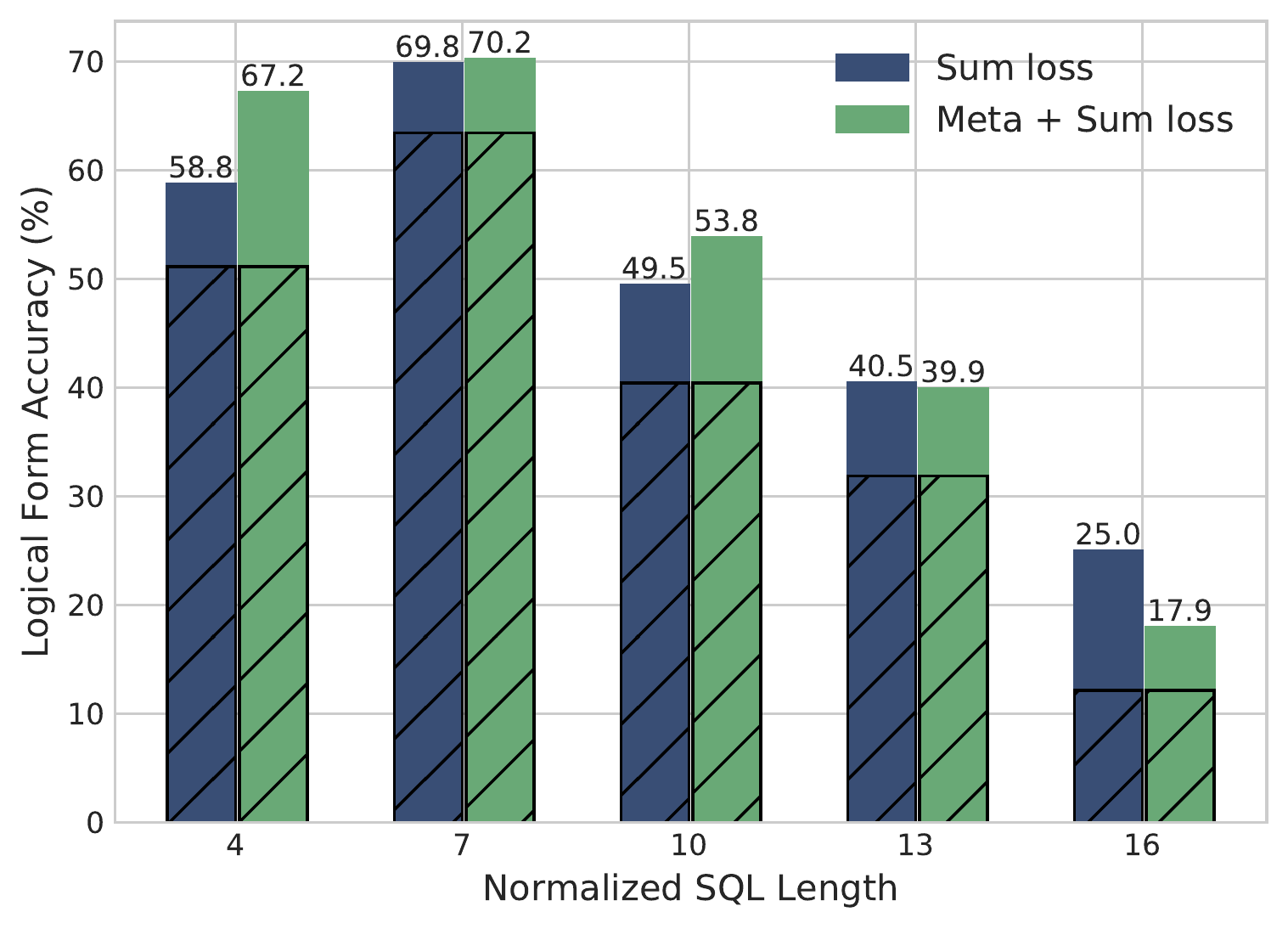}
	\vspace{-3mm}
	\caption{
		Logical form accuracy comparison for different normalized SQL lengths, where the hashed areas indicate the examples for which both models predict correctly. Note that ``Normalized SQL Length'' is used, with table entries and columns preprocessed as described in Section \ref{sec:dataset}.}
	\label{fig:error_comparison}
	\vspace{-0.3cm}
\end{figure}

We compare the logical form accuracy on the test set between the ``Sum loss'' model \citep{chenglong} and the proposed ``Meta + Sum loss'' model.
Among the 17,283 test examples, 6,661 and 6,428 errors are made by the ``Sum loss'' and ``Meta + Sum loss'' models respectively, while 5,190 errors are made by both models.
Figure \ref{fig:error_comparison} compares the logical form accuracy of the two models for different normalized SQL lengths.
We observe that the differences are mainly in ground truth SQL length = 4 and 10, where the ``Meta + Sum loss'' model outperforms the ``Sum loss'' model by a large margin.
Examples of the two cases are shown in Appendix A. 


%% file: appendix.tex
\clearpage
\appendix

\section{Error Analysis}
\label{sec:error_analysis}
We compare the logical form error on the test set between the ``Sum loss'' model \citep{chenglong} and the proposed ``Meta + Sum loss'' model.
\ignore{
Among the 17,283 test examples, there are 6,661 and 6,428 errors by the ``Sum loss'' and the ``Meta + Sum loss'', respectively.
There are 5,190 common errors by both models. 
We examine the test examples where ``Sum loss'' is correct while ``Meta + Sum loss'' is not and vice versa, shown in Figure \ref{fig:error_comparison}.
We observe that the differences are mainly in ground truth SQL length = 7 and 10, where the ``Meta + Sum loss'' model outperforms ``Sum loss'' model by a large margin.}
We show some examples for the two cases below. \\

\ignore{
\begin{figure}[h]
	\centering
	\includegraphics[width=0.9\columnwidth]{error_comparison.pdf}
	\vspace{-4mm}
	\caption{{\small Logical form accuracy comparison, where ``Meta + Sum loss (o), Sum loss (x)'' indicates the generated SQL is incorrect by the ``Sum loss'' model and is correct by the ``Meta + Sum loss'' model. Similarly ``Meta + Sum loss (x), Sum loss (o)'' indicates the generated SQL is incorrect by the ``Meta + Sum loss'' model and is correct by the ``Sum loss'' model.}}
	\label{fig:error_comparison}
	\vspace{-0.3cm}
\end{figure}
}


\subsection{Meta + Sum loss is correct and Sum loss is incorrect}

\paragraph{Example 1:}
{\small 
	\begin{itemize}
	\item Table: 2-17982145-1, Header: [benalla dfl, wins, losses, draws, byes, against] \vspace{-0.2cm}
	\item Question: when benall dfl is benalla dfl goorambat with less than 13 wins , what is the least amount of losses ?\vspace{-0.2cm}
	\item Ground Truth: SELECT MIN(losses) FROM 2-17982145-1 WHERE benalla dfl = goorambat AND wins $<$ 13 \vspace{-0.2cm}
	\item {\it Prediction (Sum loss): SELECT MIN(losses) FROM 2-17982145-1 WHERE wins = goorambat AND wins $<$ 13} \vspace{-0.2cm}

	\item Support 1 Table: 2-15582870-3, Header: [week, date, time (cst), opponent, result, game site, record, nfl recap] \vspace{-0.2cm}
	\item Support 1 Question: what is the lowest week that has 7:15 p.m. as the time (cst) and fedexfield as the game site ? \vspace{-0.2cm}
	\item Support 1 Ground Truth: SELECT MIN(week) FROM 2-15582870-3 WHERE time (cst) = 7:15 p.m. AND game site = fedexfield \vspace{-0.2cm}
	\item Support 2 Table: 	2-15389424-1, Header: [position, team, points, played, drawn, lost, against, difference] \vspace{-0.2cm}
	\item Support 2 Question: what is the lowest number of played of the team with 18 points and a position greater than 5 ? \vspace{-0.2cm}
	\item Support 2	Ground Truth: SELECT MIN(played) FROM 2-15389424-1 WHERE points = 18 AND position $>$ 5  \vspace{-0.2cm}
	\item {\it Prediction (Meta + Sum loss): SELECT MIN(losses) FROM 2-17982145-1 WHERE benalla dfl = goorambat AND wins $<$ 13}
\end{itemize}
}

\paragraph{Example 2:}
{\small 
	\begin{itemize}
		\item Table: 2-12207717-4, Header: [date, opponent, score, loss, record] \vspace{-0.2cm}
		\item Question: what was the loss of the game against the opponent angels with a 26-30 record ?\vspace{-0.2cm}
		\item Ground Truth: SELECT loss FROM 2-12207717-4 WHERE opponent = angels AND record = 26-30 \vspace{-0.2cm}
		\item {\it Prediction (Sum loss): SELECT loss FROM 2-12207717-4 WHERE opponent = 26-30 \vspace{-0.2cm}}
		\item Support 1 Table: 2-12475284-6, Header: [place, player, country, score, to par] \vspace{-0.2cm}
		\item Support 1 Question: what is the to par of the player from the country united states with a t5 place ? \vspace{-0.2cm}
		\item Support 1 Ground Truth: SELECT to par FROM 2-12475284-6 WHERE country = united states AND place = t5 \vspace{-0.2cm}
		\item Support 2 Table: 	2-17430107-2, Header: [res., record, opponent, method, event, round, location] \vspace{-0.2cm}
		\item Support 2 Question: what was the result of the fight when tetsuji kato 's record was 19-9 ? \vspace{-0.2cm}
		\item Support 2 Ground Truth: SELECT res. FROM 2-17430107-2 WHERE record = 19-9 \vspace{-0.2cm}
		\item {\it Prediction (Meta + Sum loss): SELECT loss FROM 2-12207717-4 WHERE opponent = angels AND record = 26-30}
	\end{itemize}
}

\paragraph{Example 3:}
{\small 
	\begin{itemize}
		\item Table: 1-26223231-1, Header: [season, series, team, races, wins, poles, f/laps, podiums, points, position] \vspace{-0.2cm}
		\item Question: when there are 2 wins , how many poles are ? \vspace{-0.2cm}
		\item Ground Truth: SELECT poles FROM 1-26223231-1 WHERE wins = 2 \vspace{-0.2cm}
		\item {\it Prediction (Sum loss): SELECT COUNT(poles) FROM 1-26223231-1 WHERE wins = 2}\vspace{-0.2cm}
		
		\item Support 1 Table: 2-16353025-1, Header: [week, date, opponent, result, tv time] \vspace{-0.2cm}
		\item Support 1 Question: november 25, 2001 was what week of the season ? \vspace{-0.2cm}
		\item Support 1 Ground Truth: SELECT COUNT(week) FROM 2-16353025-1 WHERE date = november 25, 2001 \vspace{-0.2cm}
		\item Support 2 Table: 	1-23285805-5, Header: [game, date, team, score, high points, high rebounds, high assists, location attendance, record] \vspace{-0.2cm}
		\item Support 2 Question: name the number of record for 30 game \vspace{-0.2cm}
		\item Support 2 Ground Truth: SELECT COUNT(record) FROM 1-23285805-5 WHERE game = 30 \vspace{-0.2cm}
		\item {\it Prediction (Meta + Sum loss): SELECT poles FROM 1-26223231-1 WHERE wins = 2}
	\end{itemize}
}

\subsection{Meta + Sum loss is incorrect and Sum loss is correct}
\paragraph{Example 4:}
{\small 
	\begin{itemize}
		\item Table: 1-22546460-4 , Header: [best male mc, best female mc, best male artist, best female artist, best male lyricist, best female lyricist, best male record] \vspace{-0.2cm}
		\item Question: who won the best female artist where best r\&b contributor won the best male lyricist ? \vspace{-0.2cm}
		\item Ground Truth: SELECT best female artist FROM 1-22546460-4 WHERE best male lyricist = best r\&b contributor \vspace{-0.2cm}
		\item {\it Prediction (Sum loss): SELECT best female artist FROM 1-22546460-4 WHERE best male lyricist = best r\&b contributor} \vspace{-0.2cm}
		
%
		\item Support 1 Table: 2-13663314-1, Header: [week, date, opponent, result, tv time] 
		\item Support 1 Question: november 25, 2001 was what week of the season ? \vspace{-0.2cm}
		\item Support 1 Ground Truth: SELECT COUNT(week) FROM 2-13663314-1 WHERE date = november 25, 2001 \vspace{-0.2cm}
		\item Support 2 Table: 	2-15122771-1, Header: [round, pick, player, position, school/club team] \vspace{-0.2cm}
		\item Support 2 Question: what is round 1 's position ? \vspace{-0.2cm}
		\item Support 2 Ground Truth: SELECT position FROM 2-15122771-1 WHERE round = 1 \vspace{-0.2cm}
		\item {\it Prediction (Meta + Sum loss): SELECT best female artist FROM 1-22546460-4 WHERE best male lyricist = best male lyricist}
	\end{itemize}
}

\paragraph{Example 5:}
{\small 
	\begin{itemize}
		\item Table: 2-1014145-1 , Header: [pick \#, mls team, player, position, affiliation] \vspace{-0.2cm}
		\item Question: tell me the lowest pick number for columbus crew \vspace{-0.2cm}
		\item Ground Truth: SELECT MIN(pick \#) FROM 2-1014145-1 WHERE mls team = columbus crew \vspace{-0.2cm}
		\item {\it Prediction (Sum loss): SELECT MIN(pick \#) FROM 2-1014145-1 WHERE mls team = columbus crew } \vspace{-0.2cm}
		
		%
		\item Support 1 Table: 1-184334-2, Header: [district, s barangay, population (2010 census), area (has .), pop. density (per km2)] \vspace{-0.2cm}
		\item Support 1 Question: what is the population (2010 census) if the area is area (has .) 66.11 ? \vspace{-0.2cm}
		\item Support 1 Ground Truth: SELECT MIN(population (2010 census)) FROM 1-184334-2 WHERE area (has .) = 66.11 \vspace{-0.2cm}
		\item Support 2 Table: 	1-22402438-7, Header: [pick \#, player, position, nationality, nhl team, college/junior/club team] \vspace{-0.2cm}
		\item Support 2 Question: how many pick \# are there for the goaltender position ? \vspace{-0.2cm}
		\item Support 2 Ground Truth: SELECT MIN(pick \#) FROM 1-22402438-7 WHERE position = goaltender \vspace{-0.2cm}
		\item {\it Prediction (Meta + Sum loss): SELECT MIN(pick \#) FROM 2-1014145-1 WHERE pick \# = columbus crew}
	\end{itemize}
}

\paragraph{Example 6:}
{\small 
	\begin{itemize}

		\item Table: 2-18391739-1  , Header: [year, stage, start of stage, distance (km), category of climb, stage winner, yellow jersey] \vspace{-0.2cm}
		\item Question: what is the distance for stage winner jos\'e-manuel fuente when the stage was less than 16 ? \vspace{-0.2cm}
		\item Ground Truth: SELECT distance (km) FROM 2-18391739-1 WHERE stage $<$ 16 AND stage winner = jos\'e-manuel fuente \vspace{-0.2cm}
		\item {\it Prediction (Sum loss): SELECT distance (km) FROM 2-18391739-1 WHERE stage $<$ 16 AND stage winner = jos\'e-manuel fuente} \vspace{-0.2cm}
		
		\item Support 1 Table: 2-1676921-5 , Header: [date, tournament, surface, partner, opponents in final, score in final] \vspace{-0.2cm}
		\item Support 1 Question:  is the opponents in final in the match with a score in final of 4--6, 1--6 played on clay surface ? \vspace{-0.2cm}
		\item Support 1 Ground Truth: SELECT opponents in final FROM 2-1676921-5 WHERE score in final = 4--6, 1--6 AND surface = clay \vspace{-0.1cm}
		\item Support 2 Table: 	1-170958-2, Header: [official name, status, area km 2, population, census ranking] \vspace{-0.2cm}
		\item Support 2 Question: what is the land area of official name hopewell parish in km2 ? \vspace{-0.2cm}
		\item Support 2 Ground Truth: SELECT area km 2 FROM 1-170958-2 WHERE official name = hopewell \vspace{-0.2cm}
		\item {\it Prediction (Meta + Sum loss): SELECT distance (km) FROM 2-18391739-1 WHERE stage winner = jos\'e-manuel fuente AND stage $<$ 16}
	\end{itemize}
}